\RequirePackage[2020-02-02]{latexrelease}

\documentclass[10pt,twocolumn,letterpaper]{article}
\usepackage{preprint}

\usepackage{amsmath, amsthm, amssymb, amsfonts}

\usepackage[numbers,square]{natbib}
\usepackage[utf8]{inputenc}	
\usepackage[T1]{fontenc}	
\usepackage{xcolor}		
\usepackage[colorlinks = true,
            linkcolor = purple,
            urlcolor  = blue,
            citecolor = cyan,
            anchorcolor = black]{hyperref}	
            
\usepackage{booktabs} 		
\usepackage{nicefrac}		
\usepackage{microtype}		
\usepackage{lineno}		
\usepackage{float}			

\usepackage{lipsum}		

\usepackage{newfloat}
\DeclareFloatingEnvironment[name={Supplementary Figure}]{suppfigure}
\usepackage{sidecap}
\sidecaptionvpos{figure}{c}

\usepackage{titlesec}
\titlespacing\section{0pt}{12pt plus 3pt minus 3pt}{1pt plus 1pt minus 1pt}
\titlespacing\subsection{0pt}{10pt plus 3pt minus 3pt}{1pt plus 1pt minus 1pt}
\titlespacing\subsubsection{0pt}{8pt plus 3pt minus 3pt}{1pt plus 1pt minus 1pt}

\usepackage{tikz,xcolor,hyperref}

\usepackage{graphicx}
\usepackage{amsmath}
\usepackage{amssymb}
\usepackage{booktabs}
\usepackage{bm}
\usepackage{setspace}
\usepackage{color}
\usepackage{arydshln}
\usepackage{algorithmic}
\usepackage{algorithm}
\usepackage{graphicx}
\usepackage{amssymb, amsmath}
\usepackage{booktabs}
\usepackage{multirow}
\usepackage{amsmath}
\usepackage{accents}
\usepackage{subcaption}

\usepackage[capitalize]{cleveref}
\crefname{section}{Sec.}{Secs.}
\Crefname{section}{Section}{Sections}
\Crefname{table}{Table}{Tables}
\crefname{table}{Tab.}{Tabs.}

\definecolor{lime}{HTML}{A6CE39}
\DeclareRobustCommand{\orcidicon}{
	\begin{tikzpicture}
	\draw[lime, fill=lime] (0,0) 
	circle [radius=0.16] 
	node[white] {{\fontfamily{qag}\selectfont \tiny ID}};
	\draw[white, fill=white] (-0.0625,0.095) 
	circle [radius=0.007];
	\end{tikzpicture}
	\hspace{-2mm}
}
\foreach \x in {A, ..., Z}{\expandafter\xdef\csname orcid\x\endcsname{\noexpand\href{https://orcid.org/\csname orcidauthor\x\endcsname}
			{\noexpand\orcidicon}}
}

\title{The effects of short video-sharing services on video copy detection}

\usepackage{authblk}

\author[1]{Rintaro Yanagi\orcidA{}}
\author[2]{Yamato Okamoto\orcidB{}}
\author[3]{Shuhei Yokoo}
\author[4]{Shin’ichi Satoh\orcidD{}}

\affil[1]{Hokkaido Univ}
\affil[2]{LINE WORKS Corp}
\affil[3]{LY Corporation}
\affil[4]{National Institute of Informatics}

\begin{document}

\twocolumn[ 
  \begin{@twocolumnfalse} 
  
\maketitle

\begin{abstract}
    The short video-sharing services that allow users to post 10-30 second videos (e.g., YouTube Shorts and TikTok) have attracted a lot of attention in recent years.
    However, conventional video copy detection (VCD) methods mainly focus on general video-sharing services (e.g., YouTube and Bilibili), and the effects of short video-sharing services on video copy detection are still unclear. 
    Considering that illegally copied videos in short video-sharing services have service-distinctive characteristics, especially in those time lengths, the pros and cons of VCD in those services are required to be analyzed. 
    In this paper, we examine the effects of short video-sharing services on VCD by constructing a dataset that has short video-sharing service characteristics.
    Our novel dataset is automatically constructed from the publicly available dataset to have reference videos and fixed short-time-length query videos, and such automation procedures assure the reproducibility and data privacy preservation of this paper.
    From the experimental results focusing on segment-level and video-level situations, we can see that three effects: ``Segment-level VCD in short video-sharing services is more difficult than those in general video-sharing services'', ``Video-level VCD in short video-sharing services is easier than those in general video-sharing services'', ``The video alignment component mainly suppress the detection performance in short video-sharing services''.
\end{abstract}
\vspace{0.35cm}

  \end{@twocolumnfalse} 
] 



\section{Introduction}
%
%
%
%
Short video-sharing services that allow users to post 10-30 second videos (e.g., YouTube Shorts, TikTok, and Instagram reels) have been attracting a lot of attention with the recent spread of Web and streaming technologies~\cite{xiao2019research,han2022research,chen2019study,he2020liveclip}.
In such video-sharing services, video copy detection (VCD) is a fundamental system for constructing fair and safe services, and accurately detecting illegally copied videos can protect video creators and ensure the reliability of each service~\cite{liu2013near,shelke2021comprehensive}.
However, the main topic of conventional research is VCD in general video-sharing services, and the effects of short video-sharing services on VCD have not been sufficiently examined.
For constructing safe and reliable short video-sharing services, the effects of short video-sharing services on VCD are desirable to be confirmed.\par
\begin{figure}[t]	
  \centerline{\includegraphics[scale=0.38]{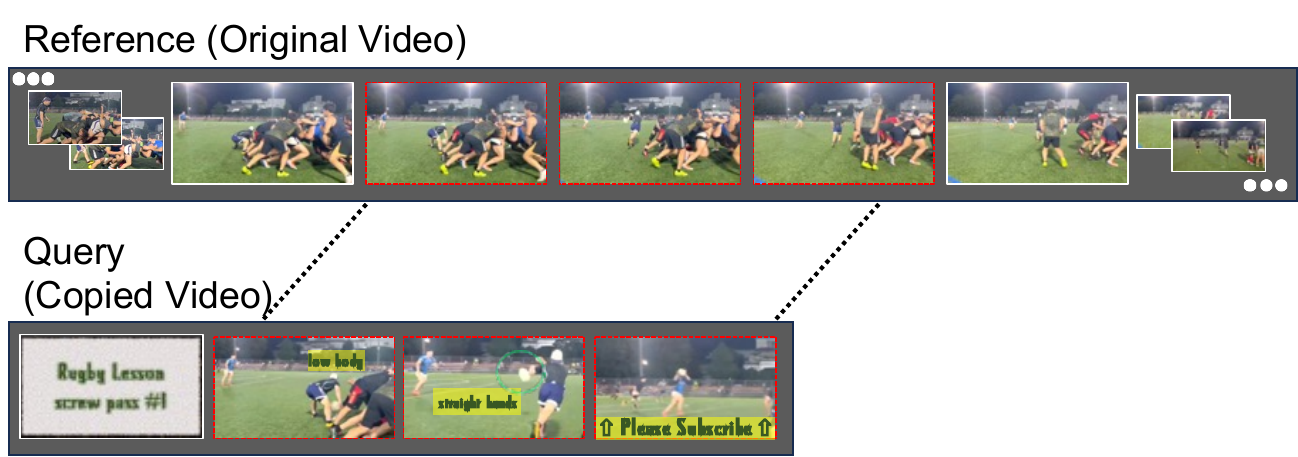}}
  \vspace{5pt}
  \caption{The brief overview of the VCD in this paper. In VCD, query video is created from the part of the original reference videos by applying various edits. By detecting the query video and copy region based on the reference videos, illegally reproduced videos can be accurately detected for realizing safe and reliable short video-sharing services.}\medskip
  \label{fig:copy_sample}
    \vspace{5pt}
\end{figure} 
The distinctive characteristic of VCD in short video-sharing services is the asymmetry of video time length. 
Focusing on video time lengths in short video-sharing services, the detection target videos (query videos) are significantly shorter than the original source videos (reference videos) as shown in \cref{fig:copy_sample}.
%
%
Actually, while the time lengths of posted videos are limited to about 10-30 seconds in Youtube Shorts\footnote{https://support.google.com/youtube/answer/10059070?hl=en}, TikTok\footnote{ https://www.tiktok.com/en}, and Instagram reels\footnote{https://about.instagram.com/features/reels}, the time lengths of reference videos are more than 60 seconds because these videos are collected from TV and video-sharing services without time restriction.
Namely, for confirming the effects of short video-sharing services on VCD, validating the versatility and applicability of conventional VCD methods to video time length asymmetry is important.
\par
\begin{figure}[!t]
    \centering
    \begin{minipage}[t]{0.48\linewidth}
        \centering
        \includegraphics[width=\linewidth]{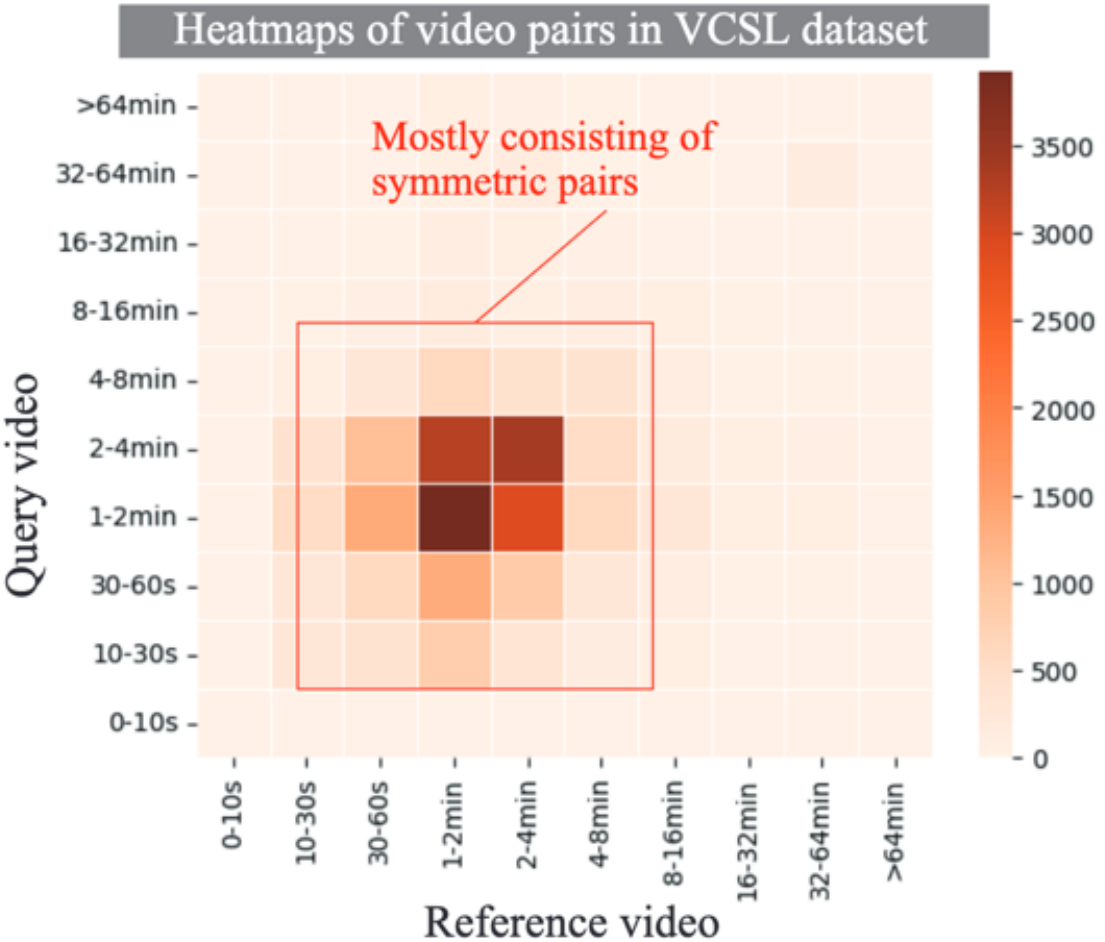}
        \subcaption{The VCSL}
        \label{fig:heat1}
    \end{minipage}
    \hfill
    \begin{minipage}[t]{0.48\linewidth}
        \centering
        \includegraphics[width=\linewidth]{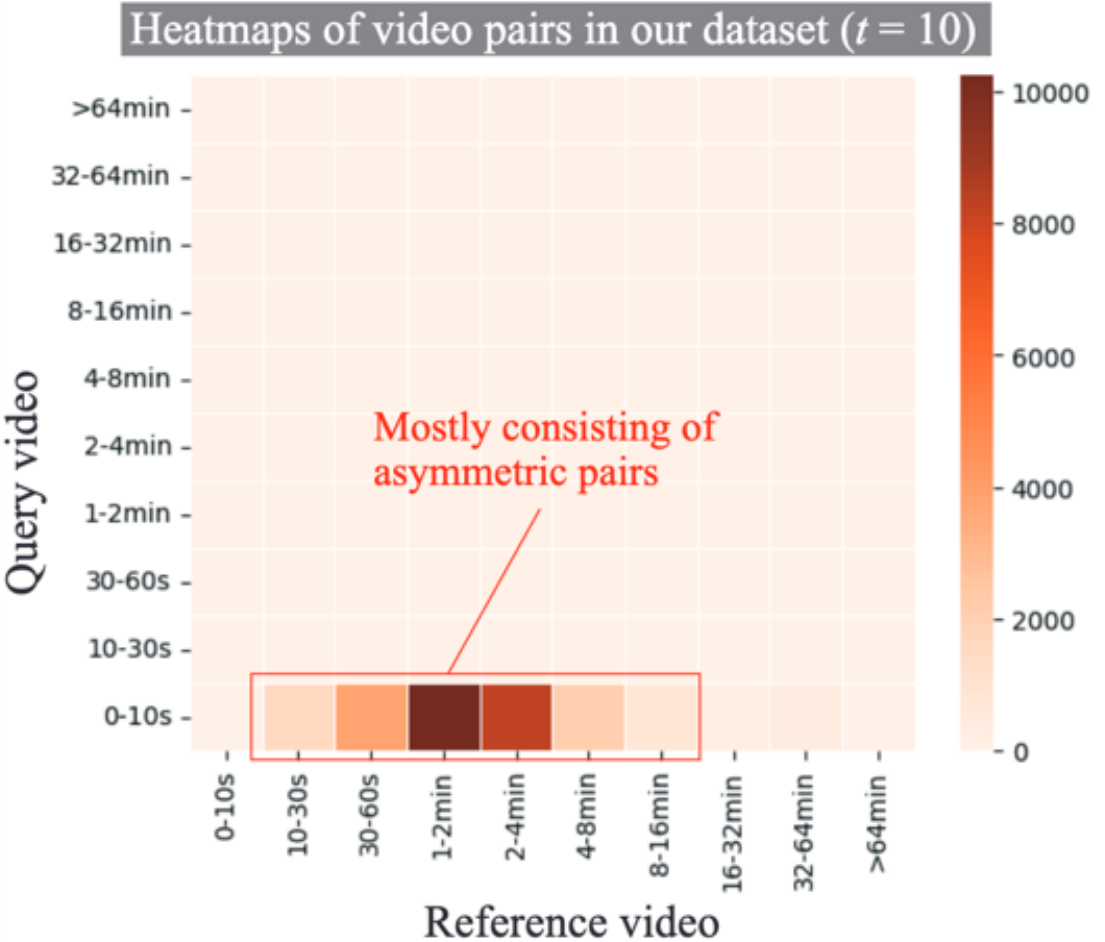}
        \subcaption{Our Dataset}
        \label{fig:heat2}
    \end{minipage}
    \caption{The video time-length heatmap of reference-query video pairs in the VCSL dataset and our dataset ($t=10$). While the most of reference-query video pairs in the VCSL dataset are symmetric in time length, query videos are significantly shorter than reference videos in our dataset. As a result, our dataset offers a more practical evaluation of VCD performance under short video-sharing service situations.}
    \label{fig:heat}
    \vspace{10pt}
\end{figure}

For evaluating the effectiveness of conventional VCD methods to video time length asymmetry, it is required to prepare an asymmetric dataset that contains query and reference videos with long and short time lengths, respectively.
However, since videos are collected from TV and video-sharing services without time restriction, the lengths of query and reference videos in conventional public VCD datasets~\cite{he2022transvcl,he2022large,jiang2014vcdb,wu2007practical,kordopatis2019fivr,law2007muscle} are generally not asymmetric as shown in \cref{fig:heat1}.
Therefore, even if VCD methods are effective in the conventional public VCD dataset, their effectiveness in short video-sharing services is not yet clear.
Since clues for VCD in short videos are short along with the time length of those videos, VCD in short video-sharing services may be a more challenging task than in general video-sharing services.
Considering that the market for short video-sharing services is expanding, confirming the effectiveness of VCD methods under the asymmetric short video-sharing service situation is desirable.\par
\newpage
In this paper, we examine the effects of VCD, especially focusing on the current mainstream short video-sharing services. 
%
To the best of our knowledge, there are no publicity available datasets focusing on the characteristics of short video-sharing services.
Therefore, in our paper, we reconstruct the most popular public dataset called VCSL~\cite{he2022large} by assuming the characteristics of short video-sharing services, and confirm the detection performance in the reconstructed dataset. 
In our experiments, we find that the performance of conventional VCD methods under short video-sharing services situations is decreased and increased at the segment level and at the video level, respectively.
These results provide insights for realizing safe and reliable short video-sharing services.
Furthermore, to analyze the core weak points of VCD methods and indicate future directions, we conducted the bottleneck analysis using the ideal feature extractor constructed based on ground truth information. 
Our experimental results imply that video alignment in the conventional VCD methods is the primary bottleneck under short video-sharing service situations, and the refinement of video alignment is beneficial to VCD in short video-sharing services.\par
The contributions of this paper are summarized as follows:
\begin{description}
\item[\textbf{A dataset for short video-sharing services}]~\\
For examining the effects of short video-sharing services on VCD, we construct a dataset containing various time-length reference videos and fixed short-time-length query videos.
\item[\textbf{Comprehensive examination on various methods}]~\\
By using the reconstructed dataset, we confirm the performance of the conventional VCD methods via segment-level, video-level, and component analysis experiments. These results provide insights for realizing safe and reliable short video-sharing services.
\end{description}

\section{Related works}
\begin{table*}[!t]
\caption{Comparison between our dataset and existing datasets: Our dataset is reconstructed from the VCSL, a realistic copied dataset, so its type is defined as "semi-simulated". Its number of labeled videos is the same as that in the VCSL. A uniqueness of our dataset is that to simulate short video-sharing services, the query videos in our dataset are significantly shorter than the reference videos, i.e., asymmetry.
} 
\vspace{10pt}
\centering
\small
\setlength\tabcolsep{3.9pt}
\begin{tabular}{cccccccccc}
\hline
Item & CCWEB & MUSCLE-VCD & TRECVID & FIVR & SVD & FIVR-PVCD & VCDB & VCSL & Ours\\
\hline
Types of copies & Realistic & Simulated & Simulated & Realistic  &  Realistic & Realistic & Realistic & Realistic & Semi-Simulated \\
\#labeld videos & 12,790 &  101 & 11,503 & 12,868 & 34,020 & 5,964 & 528 & 9,207 & 9,207 \\
 Video length & Various & Various & Various & Various & Various & Various & Various & Various & Asymmetry \\
\hline
 \end{tabular}
 \normalsize
\label{Table:dataset}
  \vspace{20pt}

\end{table*}  
\subsection{Methods for video copy detection}
The aim of VCD is to find copy videos from a given query video, and its task can be roughly classified into two categories: video-level VCD and segment-level VCD. Video-level VCD ranks copy videos ahead of irrelevant ones while segment-level VCD seeks the accurate corresponding copied segments from a given query and target video. Both video-level and segment-level VCD are essential for detecting copied videos in video-sharing services since the former and the latter contribute the computational efficiency and detection accuracy.\par
The conventional video-level VCD methods are generally realized by extracting global video-level features, calculating similarities between the extracted features, and ranking candidate videos based on the calculated similarities~\cite{wu2007practical,huang2010practical,gao2017er3,kordopatis2017near,lee2018collaborative,lee2020large}. These approaches are also known as coarse-grained approaches, and the main concern is how to calculate the global video-level features. Early works extracted hand-crafted features (e.g., color histograms) from each video frame and aggregated them into a global video-level feature~\cite{wu2007practical,huang2010practical}. With the recent progress of neural network technologies, recent works rely on CNN and Transformer-based feature extractors combined with aggregation methods~\cite{gao2017er3,kordopatis2017near,lee2018collaborative,lee2020large}. By using the video-level features for detecting copy videos, coarse-grained approaches can provide very efficient retrieval speed and contribute to the scalability of large-scale applications. \par
Although video-level VCD methods are helpful for realizing computationally efficient detection, the performance of these approaches is limited since these methods do not focus on frame-level similarity.
For further improving detection performance, segment-level VCD methods that use frame-level features have been proposed~\cite{douze2010image,tan2009scalable,berndt1994using,chou2015pattern,jiang2021learning}. These approaches are also called fine-grained approaches, and the general process of these approaches is composed of two steps: visual feature extraction and video temporal alignment. Most of the conventional methods focus on the step of visual feature extraction, and the most representative features extraction method is the ISC-based method that uses data augmentation, contrastive learning, and progressive learning for the copy-detection task~\cite{yokoo2021contrastive}. As the step of the video temporal alignment, various traditional methods are applied such as Temporal Network~\cite{tan2009scalable} and Dynamic Programming~\cite{chou2015pattern}. The temporal Network detects the longest shared path between two compared videos based on the extracted frame-level features, and Dynamic Programming extracts the diagonal blocks with the largest similarity based on the similarity matrix between frame-level features. As the most recent work, Similarity Pattern Detection (SPD) simply applies the object detection module that is trained to detect copy regions from the similarity matrix~\cite{jiang2021learning}, and it achieves better detection performance. \par
Although various approaches have been proposed to improve both computational efficiency and detection accuracy, the effectiveness of these approaches under short video-sharing service situations has not been evaluated. By evaluating their performances in this paper, we contribute to the construction of fair and safe short video-sharing services.

\vspace{15pt}
\subsection{Dataset for video copy detection}
Various datasets have been widely proposed for VCD. These datasets can be classified based on the annotation granularity: video-level and segment-level annotations. CCWEB~\cite{wu2007practical}, FIVR~\cite{kordopatis2019fivr}, and SVD~\cite{jiang2019svd} are representative datasets with video-level annotation, and video-level annotation only indicates whether two videos contain copied parts or not. Although these datasets with video-level annotation are useful for evaluating video-level VCD performance, it is difficult to evaluate segment-level VCD performance. For the segment-level evaluation, early works try to produce automatically segment-level labels by generating simulated copied segments with pre-defined transformations e.g., MUSCLE-VCD~\cite{law2007muscle} and TRECVID~\cite{kraaij2011trecvid}. Recently, VCDB~\cite{jiang2014vcdb} which is a manually-labeled segment-level dataset has been widely used for evaluating the performance of video copy detection. Although the VCDB dataset is the first dataset containing manually-labeled segment-level annotations, the size of its dataset is not large. To overcome its issue, the VCSL dataset that contains 160k infringed video pairs with 280k carefully annotated segment pairs has been proposed. Thanks to the VCSL dataset~\cite{he2022large}, segment-level video copy detection evaluation with a large-scale dataset have been realized. However, the conventional datasets are constructed from not short-video sharing services (e.g., YouTube and Bilibili), and the performance of VCD on short-video sharing services (e.g., YouTube Short, TikTok, and LINE Voom) have been not evaluated. To validate the effectiveness of the conventional VCD methods, we construct a dataset containing various time-length reference videos and fixed short-time-length query videos. The difference between our dataset and the other datasets is shown in \cref{Table:dataset}. By using the constructed dataset for evaluation, insights into VCD on the short video-sharing services are provided.

\vspace{10pt}
\section{Dataset for video copy detection on short video-sharing services}
\begin{figure}[!t]	
  \centerline{\includegraphics[scale=0.82]{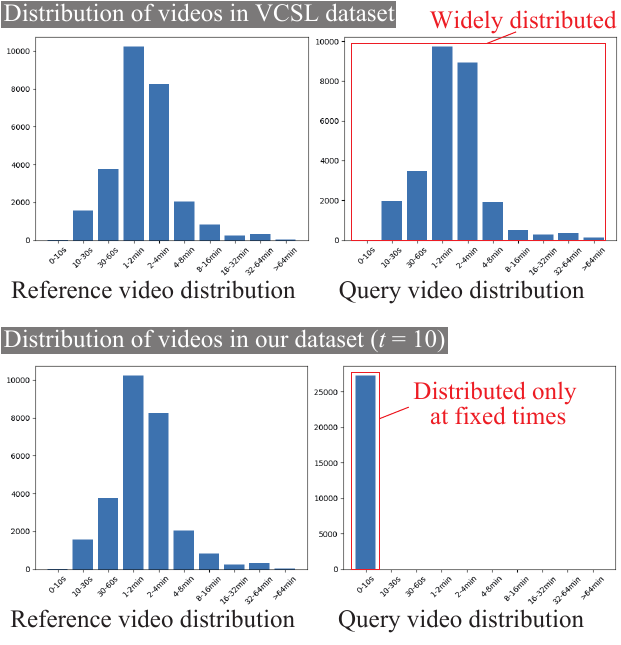}}
  \vspace{5pt}
  \caption{The time distribution of reference and query videos in VCSL dataset and our dataset ($t=10$). Different from the VCSL dataset which includes widely distributed query videos, our dataset only includes fixed-time query videos.}\medskip
  \label{fig:dist}
  \vspace{10pt}
\end{figure}

To evaluate VCD methods under short video-sharing service situations, we construct an asymmetric dataset that contains various time-length reference videos and fixed short-time-length query videos based on the VCSL dataset. 
The VCSL dataset contains a lot of reference videos and query videos, and the partial regions of the query video are manually labeled as a copy of the reference video. These video pairs are called reference-query video pairs, and each frame in reference-query video pairs is labeled as copy or not copy. The reference and query videos in VCSL dataset are collected from YouTube and Bilibili, and these videos are crawled based on 122 queries in 11 common topics, i.e., movies, TV series, music videos, sports, games, variety shows, animation, daily life, advertisement, news, and kichiku. 
%
Despite the asymmetry and shortness of video time length in short video-sharing services, in the VCSL dataset, the query and reference videos are not asymmetric, and the query videos have various time lengths (as shown in \cref{fig:heat1}). Therefore, the current VCSL dataset is not suitable for evaluating VCD methods under the short video-sharing service situation. Therefore, we modify the current VCSL dataset to have asymmetric characteristics by editing the query videos ensuring that the time length of the edited query videos is a fixed small value $t$. The time length distribution difference between VCSL and our dataset is shown in \cref{fig:dist}. It should be noted that the edited query videos definitely contain the copy region and the reference videos are used as is.\par
\begin{figure}[!t]	
  \centerline{\includegraphics[scale=0.85]{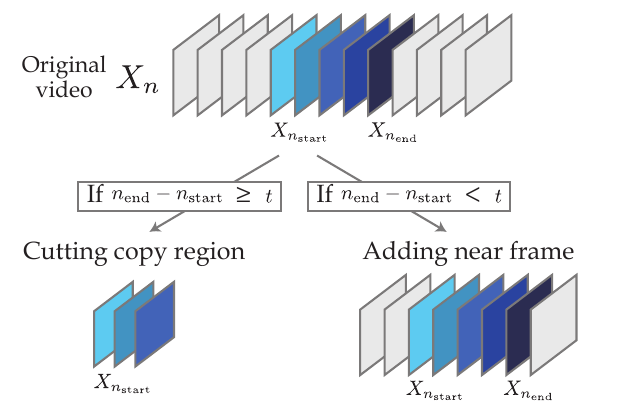}}
  \vspace{10pt}
  \caption{The overview of the dataset construction process.}\medskip
  \label{fig:overview}
\end{figure} 
\begin{algorithm}[t]
\label{algorithm}
\caption{Dataset construction}         
\begin{algorithmic}
\STATE $c = n_\text{end} - n_\text{start}$
\IF{$c \geq t$}
\FOR{$m = 1 \, \ldots \, t$}
\STATE $X^\text{edit}_m \leftarrow X_{n_\text{start} + m}$
\ENDFOR
\ENDIF
\IF{$c < t$}
\STATE $m = \min(t-c, n_\text{start})$
\STATE $p \gets \text{random}(0,m)$
\FOR{$m = 1 \, \ldots \, t$}
\STATE $X^\text{edit}_m \leftarrow X_{n_\text{start} - p + m}$
\ENDFOR
\ENDIF
\end{algorithmic}
\end{algorithm}

The basic idea of dataset construction is as follows: ``If the length of the copy region is longer than $t$, only the $t$ time-length copy region is used as a query video.'' and ``If the length of the copy region is shorter than $t$, former and latter frames of the copy region is also used as a query video.''. The overview of the dataset construction process is shown in \cref{fig:overview} and the algorithm 1. Let $X_n$ ($n = 1,\dots,N$; $N$ being the time length of the original query video) be the query video. Here, in the VCSL dataset, each video is sampled to contain one frame per second. Namely, $N$ and $X_n$ indicate the number of seconds in each video and frame of $n$-th second, respectively. By assuming the $X_{n_\text{start}}$ to $X_{n_\text{end}}$ is the copy start and end frame, the edited video $X^\text{edit}_m$ ($m = 1,\dots,t$) is constructed following the algorithm 1. In algorithm 1, $\text{random}(0,m)$ returns a uniform random number from 0 to $m$. If the length of the copy region is longer than $t$, we cut the copy region so that the time length is $t$. Following the above process, we adjust the copy region labels of the query and reference videos. If the length of the copy region is shorter than $t$, randomly selected former and latter frames are used as the query video in addition to the copy region. By randomly selecting the former and latter frames, we can create the query video without the bias of copy region location. Here, since it is difficult to create $t$ time length videos from videos less than $t$ time length, the videos less than $t$ time length are excluded. By conducting the above edit for each query video in the VCSL dataset, we obtain a dataset that contains reference videos and the corresponding $t$ time-length query videos. As shown in \cref{fig:heat2}, we can see that the query video distribution in our dataset is distributed only at fixed times and the most of reference-query video pairs in our dataset are asymmetric in time length, respectively. With the above-described process, we collect 6 datasets containing specific time-length query videos by changing hyperparameter $t$ from 10 to 60 since the posted video lengths are limited to about 60 seconds in short video-sharing services. The positive and negative samples of each dataset are both 27,765 samples. By evaluating VCD methods on the reconstructed dataset, we confirm the effects of short video-sharing services on VCD.


\vspace{10pt}
\section{Experiments of segment-level VCD}\label{Sec.segment-level}
\vspace{10pt}
Both segment-level and video-level VCD are essential in video-sharing services since the former and the latter contribute to detection accuracy and computational efficiency, we evaluated both VCD performances on the reconstructed dataset. At first, to evaluate the segment-level VCD performance on short video-sharing services, we evaluated the localization performance of the conventional fine-grained VCD methods using the reconstructed dataset. We describe the experimental settings and discuss the obtained experimental results.

\begin{table*}[!t]
\caption{Experimental results of the reconstructed dataset with each $t$ and the original VCSL using segment-level evaluation metrics. } 
\centering
\small
  \vspace{5pt}
\scalebox{1.13}{
\begin{tabular}{lcccccccc}
\hline
 method & metrics & $t$=10  & $t$=20 & $t$=30  & $t$=40 & $t$=50 & $t$=60 & VCSL  \\
 \hline\hline
  \multirow{3}{*}{HV} & SR & 14.53 & 17.08 & 21.22 & 26.73 & 32.22 & 37.41 & 40.11 \\
                      & SP & 48.77 & 47.03 & 46.98 & 46.77 & 46.21 & 46.01  & 58.77 \\
                      & SF1 & 22.39 & 25.05 & 29.23 & 34.02 & 37.97 & 41.26 & 47.67\\
                      \hline
 \multirow{3}{*}{TN} & SR & 52.24 & 55.74 & 60.01 & 63.41 & 64.21 & 67.65 & 71.66 \\
                      & SP & 55.73 & 54.81 & 54.70 & 54.51 & 54.44 & 54.04 & 56.31  \\
                      & SF1 & 53.93 & 55.27 & 57.23 & 58.62 & 58.92 & 60.08 & 63.07 \\
                      \hline
 \multirow{3}{*}{DP} & SR & 21.34 & 27.70 & 30.21 & 35.11 & 39.64 & 41.33 & 48.61 \\
                      & SP & 54.21 & 54.09 & 53.88 & 53.68 & 53.44 & 53.02 & 65.81 \\
                      & SF1 & 30.62 & 36.63 & 38.71 & 42.45 & 45.51 & 46.45 & 55.91 \\
                      \hline
 \multirow{3}{*}{DTW} & SR & 19.73 & 21.32 & 23.11 & 25.09 & 27.56 & 30.66 & 42.41\\
                      & SP & 50.44 & 49.91 & 49.53 & 49.01 & 48.72 & 48.50 & 61.55 \\
                      & SF1 & 28.36 & 29.87 & 31.51 & 33.18 & 35.20 & 37.56 & 50.21 \\
                      \hline
 \multirow{3}{*}{SPD} & SR & 24.47 & 33.51 & 39.67 & 41.75 & 42.01 & 44.56 & 51.48 \\
                      & SP & 56.95 & 56.83 & 56.71 & 56.66 & 56.45 & 56.31 & 72.28 \\
                      & SF1 & 34.23 & 42.16 & 46.68 & 48.07 & 48.17 & 49.75 & 60.13 \\
  \hline
 \end{tabular}
}
\label{Table:segment-level}
  \vspace{15pt}
\end{table*}  
\begin{table*}[!t]
\caption{Experimental results of the reconstructed dataset with each $t$ and the original VCSL using macro-segment-level evaluation metrics. } 
\centering
\small
\scalebox{1.13}{
  \vspace{10pt}
\begin{tabular}{lcccccccc}
\hline
 method & metrics & $t$=10  & $t$=20 & $t$=30  & $t$=40 & $t$=50 & $t$=60 & VCSL  \\
 \hline\hline
 \multirow{3}{*}{HV} & mSR & 25.11 & 27.81 & 34.40  & 41.32 & 47.66 & 50.61 & 54.32 \\
                      & mSP & 48.22 & 47.32 & 43.11 & 41.21 & 39.89 & 38.77 & 57.01 \\
                      & mSF1 & 33.02 & 35.03 & 38.26 & 41.26 & 43.43 & 43.90 & 55.63  \\
                      \hline
 \multirow{3}{*}{TN} & mSR & 52.72 & 56.04 & 60.04 & 62.13 & 64.21 & 67.54 & 86.43 \\
                      & mSP & 68.79 & 65.41 & 60.42 & 57.80 & 54.21 & 51.12 & 85.93  \\
                      & mSF1 & 59.69 & 60.36 & 60.22 & 59.88 & 58.78 & 58.19 & 86.18 \\
                      \hline
 \multirow{3}{*}{DP} & mSR & 31.44 & 37.88 & 45.31 & 49.61 & 53.20 & 57.84 & 67.50 \\
                      & mSP & 60.53 & 59.22 & 58.70 & 58.62 & 57.10 & 56.33 & 88.51 \\
                      & mSF1 & 41.38 & 46.20 & 51.14 & 53.73 & 55.18 & 57.17 & 76.59 \\
                      \hline
 \multirow{3}{*}{DTW} & mSR & 23.50 & 29.60 & 35.09 & 45.66 & 53.21 & 60.51 & 60.31 \\
                      & mSP & 67.81 & 67.41 & 67.02 & 65.99 & 65.60 & 65.55 & 85.01 \\
                      & mSF1 & 34.90 & 41.13 & 46.16 & 53.97 & 58.75 & 62.92 & 70.56 \\
                      \hline
 \multirow{3}{*}{SPD} & mSR & 25.23 & 32.63 & 40.01 & 43.63 & 55.43 & 61.43 & 80.85  \\
                      & mSP & 93.71 & 92.33 & 82.82 & 81.54 & 80.71 & 79.56 & 91.96  \\
                      & mSF1 & 39.76 & 48.21 & 53.95 & 56.84 & 65.72 & 69.32 & 86.05 \\
  \hline
 \end{tabular}
\label{Table:macro-segment-level}
}
  \vspace{15pt}
\end{table*}  
\vspace{10pt}
\subsection{Experimental settings}
\textbf{Comparative methods.} In our experiments, to confirm the effects of short video-sharing services on VCD, we evaluate 5 major VCD methods: Hough Voting (HV)~\cite{douze2010image}, Temporal Network (TN)~\cite{tan2009scalable}, Dynamic Programming (DP)~\cite{chou2015pattern}, Dynamic Time Warping (DTW)~\cite{berndt1994using}, and SPD~\cite{jiang2021learning}. Here, since the ISC-based feature extractor~\cite{yokoo2021contrastive} significantly outperforms the other extractors, we used the features extracted based on the ISC-based feature extractor in this experiment. Note that the hyperparameter of each method is tuned to achieve the best performance in 1e-2 grid search.\par
\textbf{Evaluation metrics.} In the field of VCD, segment-level and macro-segment-level evaluation metrics are used for evaluating segment-level VCD performance. The segment-level evaluation metrics are introduced with MUSCLE-VCD~\cite{law2007muscle} and VCDB datasets~\cite{jiang2014vcdb}. Among these metrics, most recent researches adopt segment-level recall (SR), segment-level precision (SP), and segment-level F1 (SF1) defined in VCDB as follows:
\begin{align}
    \text{SR} &= \frac{|\text{Correctly detected segments}|}{|\text{Ground truth copy segments}|}\\
    \text{SP} &= \frac{|\text{Correctly detected segments}|}{|\text{All detected segments}|}\\
    \text{SF1} &= \frac{2 \cdot \text{SR} \cdot \text{SP} }{\text{SR} + \text{SP}}
\end{align}
Since, in the segment-level evaluation metrics, a detected segment pair is considered correct as long as both of them have at least one frame overlap with the ground truth pair, these metrics cannot consider how much the detected region overlaps the ground truth region. To overcome these limitations, recent research~\cite{he2022transvcl,he2022large} proposes the macro-segment-level evaluation metrics: macro-segment-level recall (mSR), macro-segment-level precision (mSR), and macro-segment-level F1 (mSF1). For $X \in \{\text{reference},\text{query}\}$, each metrics is calculated as follows:
\begin{align}
    \text{mSR} &=  \prod_{X} \frac{|\text{Correctly detected } X \text{ frames}|}{|\text{Ground truth } X \text{ frames}|}&\\
        \text{mSP} &=  \prod_{X} \frac{|\text{Correctly detected } X \text{ frames}|}{|\text{All detected } X \text{ frames}|}&\\
    \text{mSF1} &= \frac{2 \cdot \text{mSR} \cdot \text{mSP} }{\text{mSR} + \text{mSP}}
\end{align}
By using these evaluation metrics, we evaluate the VCD methods and confirm the effects of short video-sharing services on VCD.

\vspace{10pt}
\subsection{Confirming the effects of short video-sharing services on segment-level VCD}\label{Subsec.seg-level}
At first, we evaluate the performance of each segment-level VCD method by using the original VCSL dataset and the reconstructed dataset with a fixed short time length. By comparing these performances, we confirm whether segment-level VCD on short video-sharing services is more difficult than those on general video-sharing services.\par
Experimental results are shown in \cref{Table:segment-level} and \cref{Table:macro-segment-level}, where $t = {10,20,30,40,50,60}$ and VCSL indicate the reconstructed dataset with $t$ time length query videos and original VCSL dataset with various time length videos, respectively. The results show that the value of SF1 and mSF1 gradually decreases as the value of $t$ decreases. Also, the value of SF1 and mSF1 on the reconstructed dataset in every $t$ are lower than those on the original VCSL dataset. From these results, we confirm that segment-level VCD assuming short video-sharing services is more difficult than those for a dataset with various time-length query videos. Specifically, from the results that the value of recall (resp. precision) gradually decreases (resp. increases) as the value of $t$ decreases, it is considered that the conventional VCD methods tend to miss the copy regions rather than falsely detect copy regions under short video-sharing services situations.


\vspace{10pt}
\subsection{Confirming the bottlenecks of video copy detection methods}\label{Subsec.bottleneck}

\begin{figure}[!t]	
  \centerline{\includegraphics[scale=0.85]{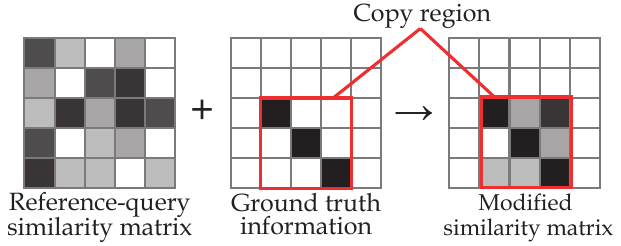}}
  \caption{The brief overview of the similarity matrix modification.}\medskip
  \label{fig:sim_mod}
    \vspace{10pt}
\end{figure} 

In \cref{Subsec.seg-level}, we confirmed that segment-level video copy detection for short query videos is more difficult than for various time-length query videos. However, it is still unclear which component of segment-level video copy detection methods adversely affects the detection performance. Next, we analyze the conventional video copy detection methods focusing on their two major components: visual feature extraction and video alignment, to obtain insights for bottleneck information. In this experiment, by refining the similarity matrix based on ground truth information, we artificially construct an ideal feature extractor that can obviously distinguish the copy and not copy region. In the experiments using the ideal feature extractor, if the visual feature extraction is the bottleneck process, the video copy detection performance is expected to be improved. Inversely, if the video alignment is the bottleneck process, its performance is expected to be not improved. Through the experiments, we confirm the bottleneck process of the current video copy detection methods.\par
The brief overview is described in \cref{fig:sim_mod}. 
At first, following the general video copy detection methods, we calculated the similarity matrix $s_{n,m}$ ($m = 1,\dots,M$; $M$ being the time length of the original query video) by extracting the features from each frame. Next, we constructed a modified similarity matrix $s^\text{mod}_{n,m}$ by assigning 0 and 1 scores for the non-target copy region in $s_{n,m}$ and the diagonal of the target copy region in $s_{n,m}$, respectively. Let $i={n_\text{cstart}},\dots,{n_\text{cend}}$ and $j = {m_\text{cstart}},\dots,{m_\text{cend}}$, where ${n_\text{cstart}}$ to ${n_\text{cend}}$ and ${m_\text{cstart}}$ to ${m_\text{cend}}$ are the copy start and end frame index. $s^\text{mod}_{n,m}$ is calculated as follows:
\begin{align}    
    s^\text{mod}_{i,j} = 
    \begin{cases}
    1 & (i-n_\text{cstart} = j - m_\text{cstart})\\
    s_{i,j} & (\text{otherwise})
    \end{cases}
\end{align}
Finally, we calculated the VCD performance using the modified similarity matrix.
\par
Experimental results are shown in \cref{Table:segment-level-gt} and \cref{Table:macro-segment-level-gt}, where methods with $\ast$ mark indicate using the above-described modified similarity matrix. 
%
On the original VCSL dataset, the values of SF1 and mSF1 were increased by using the ground truth information. These results indicate that our modified similarity matrix works well. Next, on the reconstructed $t=10$ dataset, when using modified similarity matrices, the values of precision were significantly high but the values of recall were significantly low, consequently, SF1 and mSF1 were significantly decreased. This indicates that the conventional VCD methods cannot suggest the candidate regions even when using ground truth information, and the video alignment in the conventional VCD methods suppresses the VCD performance under short video-sharing services situations. Moreover, considering that the target region in the reconstructed dataset is relatively smaller than the original VCSL dataset, the limited clue information might cause conventional video alignment methods to not work adequately.
%
Note that the hyperparameter of each method is tuned to achieve the best performance in 1e-2 grid search, and then these results do not deeply come from the hyperparameter settings. From these results, for constructing safe and reliable short video-sharing services via VCD, video alignment methods that can be used even in short video-sharing services are desirable to be considered.

\begin{table}[t!]
\caption{Experimental results for confirming the bottleneck components using segment-level evaluation metrics.} 
\vspace{5pt}
\centering
\small
\setlength\tabcolsep{4pt}
\scalebox{1.13}{
  \vspace{15pt}
\begin{tabular}{lccc|ccc}
\hline
 \multirow{2}{*}{Method} & \multicolumn{3}{c}{VCSL} & \multicolumn{3}{c}{$t$=10} \\
 & SR & SP & SF1 & SR & SP & SF1  \\
 \hline\hline
 HV & 40.11 & 58.77 & 47.67 & 14.53 & 48.77 & 22.39 \\
 HV* & 65.77 & 94.21 & 77.46 & 2.32 & 95.64 & 4.53 \\
 \hline
 TN & 71.66 & 56.31 & 63.07 & 52.24 & 55.73 & 53.93 \\
 TN*  & 75.38 & 85.25 & 80.01 & 3.43 & 98.75 & 6.62 \\
 \hline
 DP & 48.61 & 65.81 & 55.91 & 21.34 & 54.21 & 30.62 \\
 DP*  & 54.21 & 93.13 & 68.52 & 1.73 & 95.64 & 3.39 \\
 \hline
 DTW & 42.41 & 61.55 & 50.21 & 19.73 & 50.44 & 28.36 \\
 DTW*  & 45.12 & 92.42 & 60.63 & 4.78 & 91.22 & 9.08 \\
 \hline
 SPD & 51.48 & 72.28 & 60.13 & 24.47 & 56.95 & 34.23 \\
 SPD*  & 58.18 & 96.70 & 72.65 & 4.32 & 97.64 & 8.27 \\
  \hline
 \end{tabular}
\label{Table:segment-level-gt}
\vspace{15pt}
}
\vspace{15pt}
\caption{Experimental results for confirming the bottleneck components using macro-segment-level evaluation metrics.} 
\vspace{5pt}
\centering
\small
\setlength\tabcolsep{4pt}
\scalebox{1.13}{
  \vspace{15pt}
\begin{tabular}{lccc|ccc}
\hline
 \multirow{2}{*}{Method} & \multicolumn{3}{c}{VCSL} & \multicolumn{3}{c}{$t$=10} \\
 &mSR & mSP & mSF1 & mSR & mSP & mSF1  \\
 \hline\hline
 HV & 54.32 & 57.01 & 55.63 & 25.11 & 48.22 & 33.02 \\
 HV* & 60.42 & 84.52 & 70.46 & 2.63 & 95.51 & 5.11 \\
 \hline
 TN & 86.43 & 85.93 & 86.18 & 52.72 & 68.79 & 59.69 \\
 TN* & 89.36 &  93.69& 91.48 & 1.42 & 98.76 & 2.79 \\
 \hline
 DP & 67.50 & 88.51 & 76.59 & 31.44 & 60.53 & 41.38 \\
 DP* & 75.62 & 91.24 & 82.69 & 1.77 & 95.64 & 3.47 \\
 \hline
 DTW & 60.31 & 85.01 & 70.56 & 23.50 & 67.81 & 34.90 \\
 DTW* & 80.42 & 91.11 & 85.43 & 2.37 & 98.87 & 4.62 \\
 \hline
 SPD & 80.85 & 91.96 & 86.05 & 25.23 & 93.71 & 39.76 \\
 SPD* & 82.36 & 97.86 & 89.45 & 1.76 & 94.53 & 3.45 \\
  \hline
 \end{tabular}
\label{Table:macro-segment-level-gt}
\vspace{10pt}
}
\end{table} 
\begin{table*}[!t]
\caption{Experimental results of the reconstructed dataset with each $t$ and the original VCSL dataset using video-level evaluation metrics. } 
\centering
\small
  \vspace{10pt}
\scalebox{1.13}{

\begin{tabular}{lcccccccc}
\hline
 method & metrics & $t$=10  & $t$=20 & $t$=30  & $t$=40 & $t$=50 & $t$=60 & VCSL  \\
 \hline\hline
 F2F & mAP & 0.52 & 0.50 & 0.49 & 0.43 & 0.40 & 0.37 & 0.32\\
 G2G & mAP & 0.35 & 0.34 & 0.33 & 0.33 & 0.32 & 0.32 & 0.32\\
 SM2G & mAP & 0.56 & 0.55 & 0.53 & 0.52 & 0.50 & 0.49 & 0.47\\
 \hline
 \end{tabular}

\label{Table:video-level}
}
   \vspace{10pt}
\end{table*}  
\section{Experiments of video-level VCD}
In \cref{Sec.segment-level}, we focus on segment-level VCD. Next, we focus on video-level VCD. 
Specifically, we confirm whether video-level VCD in short video-sharing services is more difficult than in general video-sharing services by comparing the performance on the original VCSL dataset and our reconstructed dataset. We describe the experimental settings, and then we discuss the obtained experimental results.

\vspace{15pt}
\subsection{Experimental settings}
\textbf{Positive and negative pairs.} 
For evaluating the video-level VCD performance, we defined the video-level labels for all possible video pairs of the VCSL dataset. Since the number of test videos in the VCSL dataset is 1,099, all possible video pairs are 603,351 (= (1,099 * 1,099 - 1,099)/2 ). For all possible video pairs, we defined copy or not copy labels based on the segment-level label information of the VCSL dataset.\par
\textbf{Aggregation methods.} In this experiment, following the conventional coarse-grained VCD methods, we evaluate the performance of three baseline methods: Flame-to-Flame (F2F), Global-to-Global (G2G), and Sub-mean-to-Global (SM2G) methods. The F2F method calculates the similarities between the extracted features of each frame and assigns the maximum value of the similarities as the similarity score between videos. On the other hand, the G2G method aggregates the extracted features of all frames and calculates the similarity between the aggregated features. Here, in our paper, mean aggregation procedure is applied for the G2G method. The SM2G method is a hybrid method of the F2F and the G2G methods. The SM2G first aggregates the extracted features of several frames and calculates the similarity between the aggregated features. Finally, the SM2G assigns the maximum value of the similarities as the similarity score between videos. Here, in this experiment, we experimentally aggregate the extracted features of each 10 frames. Also, as well as \cref{Sec.segment-level}, we used the features extracted based on the ISC-based feature extractor in this experiment.\par
\textbf{Evaluation metrics.} For evaluating the video-level video copy detection performance, micro average precision (mAP)  is commonly used as an evaluation metric, and we also employ it.

\newpage
\subsection{Confirming the effects of short video-sharing services on video-level VCD}
Experimental results are shown in \cref{Table:video-level}, where $t = {10,20,30,40,50,60}$ and VCSL indicate the reconstructed dataset with $t$ time length query videos and original VCSL dataset with various time length videos, respectively. From the results, we can see that the value of mAP gradually increases as the value of $t$ decreases. From these results, we confirm that video copy detection assuming short video-sharing services is easier than those for a dataset with various time-length query videos. 

These tendencies are different from the segment-level VCD described in \cref{Sec.segment-level}. It implies that conventional VCD can detect copy videos roughly, but cannot detect copy videos in detail under short video-sharing services situations. Also, considering that each video-level VCD method can detect copy videos even though they simply compare the features extracted from each frame, the extracted features may be sufficiently useful even in the situation of short video-sharing services. 
These results further confirm that video alignment in the conventional VCD methods suppresses the detection performance on short video-sharing services as discussed in \cref{Subsec.bottleneck}.

\section{Conclusion}
In this paper, we examined the effects of short video-sharing services on video copy detection by constructing an asymmetric dataset, where copied videos are significantly shorter than original videos to simulate short video-sharing services. 
By using the reconstructed dataset, we examined segment-level and video-level VCD experiments and evaluated various conventional methods. Experimental results provide three insights about VCD on short video-sharing services: ``Segment-level VCD on short video-sharing service situation is more difficult than those in general video-sharing services'', ``Video-level VCD in short video-sharing service situation is easier than those in general video-sharing services'', and ``The video alignment in the conventional VCD methods suppress the detection performance in short video-sharing service situation''. We believe that these results cultivate VCD research for short video-sharing services. In our future work, we will construct a video alignment method that can address VCD under the situation of short video-sharing services. 
\clearpage


\normalsize
\bibliography{paper}

\end{document}